\documentclass[letterpaper, 10 pt,english]{IEEEtran}
\usepackage[T1]{fontenc}
\usepackage[latin9]{inputenc}
\usepackage{color}
\usepackage{babel}
\usepackage{amsmath}
\usepackage{amsthm}
\usepackage{amssymb}
\usepackage{graphicx}
\usepackage[unicode=true,pdfusetitle,
 bookmarks=true,bookmarksnumbered=false,bookmarksopen=false,
 breaklinks=false,pdfborder={0 0 0},pdfborderstyle={},backref=false,colorlinks=true]
 {hyperref}

\makeatletter
\theoremstyle{theorem}

\theoremstyle{remark}

\theoremstyle{definition}

\theoremstyle{assumption}

\usepackage[ruled]{algorithm2e}
\newcommand{\nosemic}{\renewcommand{\@endalgocfline}{\relax}}
\newcommand{\dosemic}{\renewcommand{\@endalgocfline}{\algocf@endline}}
\usepackage{cite}
\usepackage[paper=letterpaper,top=0.75in,bottom=0.75in,right=0.75in,left=0.75in]{geometry}
\usepackage{subfigure}
\usepackage{enumerate}

\makeatother

\newtheorem{remark}{Remark}

\newtheorem{theorem}{Theorem}
\newtheorem{assumption}{Assumption}

\providecommand{\definitionname}{Definition}
\providecommand{\remarkname}{Remark}
\providecommand{\theoremname}{Theorem}

\begin{document}
\newgeometry{top=1in,bottom=0.75in,right=0.75in,left=0.75in}
\IEEEoverridecommandlockouts 
\title{Safe Task Space Synchronization with Time-Delayed Information}
\author{Rounak Bhattacharya, Vrithik Raj Guthikonda, Ashwin P. Dani\thanks{This work was supported in part by NSF grant no. SMA-2134367 and in part by UConn OVPR's research excellence program. Rounak Bhattacharya, Vrithik Raj Guthikonda and Ashwin P. Dani are with the
Department of Electrical and Computer Engineering at University of
Connecticut, Storrs, CT 06269. Email: \{rounak.bhattacharya;vrithik.guthikonda;ashwin.dani\}@uconn.edu.}}
\maketitle
\begin{abstract} 
In this paper, an adaptive controller is designed for the synchronization of the trajectory of a robot with unknown kinematics and dynamics to that of the current human trajectory in the task space using the delayed human trajectory information. The communication time delay may be a result of various factors that arise in human-robot collaboration tasks, such as sensor processing or fusion to estimate trajectory/intent, network delays, or computational limitations. The developed adaptive controller uses Barrier Lyapunov Function (BLF) to constrain the Cartesian coordinates of the robot to ensure safety, an ICL-based adaptive law to account for the unknown kinematics, and a gradient-based adaptive law to estimate unknown dynamics. Barrier Lyapunov-Krasovskii (LK) functionals are used for the stability analysis to show that the synchronization and parameter estimation errors remain semi-globally uniformly ultimately bounded (SGUUB). The simulation results based on a human-robot synchronization scenario with time delay are provided to demonstrate the effectiveness of the designed synchronization controller with safety constraints.
\end{abstract}
\section{Introduction}
Human-robot collaborative applications often require robots to synchronize with human agents while operating in confined environments to improve the safety of humans working nearby. Many collaborative tasks can be accomplished by synchronizing the robot's end-effector motion with that of the human, e.g., collaborative displacement of a load \cite{dani2020human,ravichandar2019human}. While performing collaborative tasks, it is important to ensure that the robot's operation remains within a confined space to improve the safety of the collaboration. Human trajectories estimated using sensors may introduce time delay due to sensor processing or communication of the estimated information to the robot controller over a network. In this paper, a task space robot controller is developed to synchronize the robot's end-effector trajectories with those of the human using delayed trajectory information, while also constraining the robot's motion within prescribed bounds to ensure safe operation. 

Control of time delay systems is a well-studied topic and several important results are presented in literature on the control of time delay systems in the presence of control input delay \cite{sharma2011predictor,fischer2013saturated,mazenc2013robustness}, state delays \cite{sharma2011rise,olgac2002exact,sipahi2011stability,zhang2018adaptive}, input and state delays \cite{krstic2009delay}, and communication delay \cite{klotz2017synchronization,Nuno2011}. It is shown in \cite{olfati2004consensus} that time delay arising from information exchange between agents may lead to unstable behavior or reduced performance. In newer applications of cyber-physical-human systems \cite{annaswamy2023cyber}, time delay may be introduced by human state estimation using sensors, network communication over a channel, or human reaction to certain events. Effects of communication time delay in multi-agent robots are studied in \cite{klotz2017synchronization,liu2011synchronization}. The method in \cite{liu2011synchronization} develops a time-delayed synchronization controller using globally Lipschitz dynamics or dynamics with no drift term. The synchronization controller in \cite{klotz2017synchronization} uses one-hop delayed information from the network neighbors in a multi-agent system scenario. The synchronization controller in \cite{bhattacharya2023adaptive} is an adaptive controller for general dynamics with a fixed known time delay in communicating human state, synchronizing with respect to the delayed state. Time delay problem has also been studied in telerobotics, where delays occur during information exchange between robots, and one robot's trajectory is synchronized with time delayed trajectory of the other robot \cite{Nuno2011,liu2011controlled,hatanaka2015passivity,yousefi2017effect,yousefi2017stability,yucelen2018stability}. However, the control design does not include state constraints, which may lead to robot states leaving the safe operating region. Constrained control of input delay systems has been studied in \cite{jankovic2018control,abel2020constrained} for linear systems and for systems with multiple input delays. For state delay problems, adaptive constrained control of uncertain nonlinear systems with full state constraints has been developed in \cite{li2017adaptive} and tracking control of a robots with unknown time-varying state delays and state constraints in \cite{li2017unknowndelay}. However, the problem of communication delay with state constraints is not addressed.

The contribution of this work is to design an adaptive controller to synchronize the robot's task space trajectory with that of a human agent at current time using time delayed trajectory information in the presence of kinematic and dynamic uncertainty in the robot model. The time delay is a known bounded constant. To account for kinematic uncertainty, an Integral Concurrent Learning (ICL)-based parameter update law is used to estimate kinematic parameters, while a gradient-based update law is used to estimate parameters corresponding to uncertain robot dynamics. Using the Barrier Lyapunov function (BLF), the controller ensures that the robot's states remain within a prescribed bound, thereby ensuring safe operation during collaborative tasks. The stability analysis of the task-space synchronization of the robot's state with that of the human at current time is conducted using Lyapunov-Krasovskii (LK) functionals along with BLF, which results in semi-global uniformly ultimately bounded (SGUUB) stability of the closed-loop system. Simulation studies, conducted on a 2 DoF robot model, show that the synchronization controller achieves bounded synchronization error using time-delayed information of the human trajectory.

\section{Kinematic and Dynamic Model \label{sec:System-Modeling}}
 
\subsection{Joint Space Robot Dynamics}
The n-link robot is represented using an Euler-Lagrange system as follows
\begin{equation}
M(\theta)\ddot{\theta}+C(\theta,\dot{\theta})\dot{\theta}+f(\dot{\theta})+G(\theta)=\tau_r(t) 
\label{eq:robot_EL_eq}
\end{equation}
where $\theta(t) \in \mathbb{R}^n$, $\dot{\theta}(t) \in \mathbb{R}^n$, and $\ddot{\theta}(t) \in \mathbb{R}^n$ represent joint angles, velocities and accelerations, respectively, $\tau_r \in \mathbb{R}^n$ denotes motor torque, $M(\theta) \in \mathbb{R}^{n \times n}$, $C(\theta,\dot{\theta}) \in \mathbb{R}^{n\times n}$, $f(\dot{\theta}) \in \mathbb{R}^{n}$ are mass-inertia matrix, centripetal-Coriolis matrix, and a Linear-in-Parameters (LIP) friction, respectively, and $G(\theta) \in \mathbb{R}^{n}$ denotes gravity. Following properties are satisfied by the dynamics. \textbf{Property 1}: Matrix $M(\theta)$ is symmetric, positive definite with $\underline{m}||y_r||^2 \leq y_r^TM(\theta)y_r \leq \bar{m}||y_r||^2, \;\forall y_r \in \mathbb{R}^n$, where $\underline{m}, \bar{m} \in \mathbb{R}_+$ are known.  \textbf{Property 2}: The inertia and centripetal-Coriolis matrices satisfy the skew-symmetry property
$\bar{\eta}^T(\dot{M} - 2C)\bar{\eta} = 0, \quad \forall \bar{\eta} \in \mathbb{R}^n$. 

The human trajectories are obtained using camera sensor that provide human hand position trajectories denoted by $p(t_T)$, $\dot{p}_h(t_T)$, $\ddot{p}_{hT}$, where $t_T = t-T$ is a time delay due to image processing and/or communication \cite{trombetta2021variable}.
\begin{assumption} \label{assum:Assumption1}
It is assumed that $\theta(t)$, $\dot{\theta}(t)$, 
are measurable.
\end{assumption}
\begin{assumption} \label{assum:assumptionqhbound}
    The human hand trajectory $x_h(t) = [p_{h}^T\; \dot{p}_{h}^T\; \ddot{p}_{h}^T]^T$ is bounded and smooth such that $|p_{hi}(t)|<k_{hi},\; \forall i = \{1, 2, 3\}$, where $k_{hi}$ are known positive constants.
\end{assumption} 
\begin{remark}\label{rem:Remark1}
    Assumption \ref{assum:assumptionqhbound} implies that the human trajectories and its first and second time derivatives are bounded,  $\Vert p_h\Vert \leq p_1$, $\Vert\dot{ p}_h\Vert \leq p_2$, $\Vert\ddot{ p}_h\Vert \leq p_3$, as well as the difference of the human trajectories with the delayed human trajectories and its first and second derivatives are upper-bounded, $\Vert p_h-p_{hT}\Vert \leq \bar{p}_1$, $\Vert\dot{ p}_h-\dot{p}_{hT}\Vert \leq \bar{p}_2$, $\Vert\ddot{ p}_h-\dot{p}_{hT}\Vert \leq \bar{p}_3$.
\end{remark}
\subsection{Forward Kinematics and Velocity Jacobian}
Using forward kinematics, the robot end-effector state is related to joint angles $\theta(t)$ as, $p = h_p(\theta)$, where $p(t)= [x \; y \; z]^T \in \mathbb{R}^{n_p}$ is the Cartesian end-effector position. The time-derivative of $p(t)$ is given by
\begin{equation}
 \dot{p} = J_p(\theta) \dot{\theta}
\label{eq:der_position}
\end{equation} 
where $J_p(\theta) \in \mathbb{R}^{n_p\times n}$ is a velocity Jacobian.
\begin{assumption} \label{assum:Assumption3}
    The velocity Jacobian $J_p(\theta)$ is square, invertible, and satisfies, $\Vert J_p(\theta) \Vert \leq \bar{j}_p$ and $\Vert J_p^{-1}(\theta) \Vert \leq \bar{j}_{p_{inv}}$, and $\Vert \dot{J}_p(\theta) \Vert \leq\bar{j}_{p_{der}}\Vert\dot{\theta}\Vert$, for constants $\bar{j}_p$, $\bar{j}_{p_{inv}}$ and $\bar{j}_{p_{der}}$.
\end{assumption}
\begin{remark}
    Since $J_p(\theta)$ is square, the robot is a fully actuated robot, i.e., $n_p=n$, where $n_p=\{2,3\}$. For rest of the manuscript, $n_p=n$ is used.
\end{remark}
\noindent\textbf{Property 3}: The robot's kinematics given in (\ref{eq:der_position}) can be linearly parametrized as 
\begin{equation}
    J_p(\theta)\dot{\theta} = W_j(\theta, \dot{\theta})\zeta_j
\label{eq:kinematic_pramet}
\end{equation}
where $\zeta_j\in \mathbb{R}^{m_1}$ are unknown parameters and $W_j(\theta, \dot{\theta}) : \mathbb{R}^{n} \times \mathbb{R}^{n} \rightarrow \mathbb{R}^{n\times m_1}$ is a regressor matrix.
\section{Safe Synchronization with Time Delayed Information}
\subsection{Open Loop Error System}
The position synchronization error between the robot and the human end-effector state, $e_{p}(t) \in \mathbb{R}^{n}$, is defined as
\begin{equation}
e_p(t) = p_h(t) - p(t) 
\label{eq:pos_coord_err}
\end{equation}
Since $p_{h}(t)$ cannot be measured at current time, a measurable error, $e_{p_T}(t) \in \mathbb{R}^n$, with respect to measurable delayed human trajectory is defined as
\begin{equation}
e_{p_{T}}(t) = p_h(t_T) - p(t) 
\label{eq:pos_coord_err_T}
\end{equation}
where $p_h(t_T) = p_h(t-T) \in \mathbb{R}^n$ is the delayed human position. 
To further develop the analysis of synchronization controller for the second order E-L dynamics, an auxiliary error $\eta(t) \in \mathbb{R}^n$ is defined as
\begin{align}
\eta(t) =\hat{J}_{p}(t)^{-1} 
(\dot{p}_{h}(t) + k_1 \phi(e_p(t)) e_p(t)) - \dot{\theta}(t) 
\label{eq:eta}
\end{align}
where $k_1\in\mathbb{R}^{}$ is an auxiliary gain, $\phi(e_p): \mathbb{R}^n \rightarrow \mathbb{R}^n\times \mathbb{R}^n$ is a diagonal matrix with its elements defined as 
\begin{equation}
    \phi_{i}(e_{p_{i}}) = \frac{1}{k_{m_i}^2-e_{p_{i}}^2}, \quad\forall i \in \{1,..,n \} \label{eq:phi}
\end{equation}
such that $k_{m_{i}} \in \mathbb{R}$ are the bounds on $e_{p_{i}}$. $\hat{J_p}(\theta)$ in (\ref{eq:eta}) is an estimate of $J_p(\theta)$.
Since $\dot{p}_h(t)$ and $e_p(t)$ cannot be measured, to design synchronizing control, a measurable auxiliary error at the delayed time $\eta_T = \eta(t-T) \in \mathbb{R}^n$ is defined as
\begin{equation}
\eta_T =\hat{J}_{p}(t)^{-1} 
(\dot{p}_{h_{T}} + k_1 \phi(e_{p_{T}}) e_{p_{T}}) - \dot{\theta}(t) 
\label{eq:eta_T}
\end{equation}

Using bounds in Assumption \ref{assum:assumptionqhbound}, the desired bounds on the robot's end-effector state $k_{r_{i}}$, and using the definition of $e_p(t)$, the bounds on $ e_{p_{i}}(t)$ can be developed as $\vert e_{p_{i}} \vert <  k_{r_i} - k_{h_i} = k_{m_{i}}$, where $k_{r_i}$, $k_{m_{i}} \in \mathbb{R}^{+}$. 
To capture the safety objective, constrained sets $\mathcal{Z}_p$, and $\mathcal{Z}_i = \{ e_{p_{i}}  \in \mathbb{R} | -k_{m_{i}} < e_{p_{i}} < k_{m_{i}} \} \subset \mathbb{R}$ are defined such that $\mathcal{Z}_p \subseteq \mathcal{Z}_1 \times \mathcal{Z}_2,.., \times \mathcal{Z}_n \subset \mathbb{R}^n$ is a constrained set of synchronization error $e_p(t)$. 
\paragraph*{Safe Adaptive Synchronization Objective} The task space safe synchronization objective can be written as
\begin{align}
    \textrm{lim\:sup}_{t\rightarrow \infty} \lVert e_p(t) \rVert \leq \varepsilon_1, \; \textrm{and}\; e_p(t) \in \mathcal{Z}_p \subset \mathcal{Z} \quad \forall t\geq 0
\end{align} 
for a small $\varepsilon_1 \in \mathbb{R}^+$, $\mathcal{Z}_p$ is defined subsequently, when the parameters of the robot kinematics and dynamics are unknown.
\subsection{Error Dynamics}
Taking the time derivative of (\ref{eq:pos_coord_err}) and (\ref{eq:pos_coord_err_T}), the error dynamics can be derived as $\dot{e}_p = \dot{p}_{h} - J_p \dot{\theta}$,
\label{eq:der_tracking_error}
which can be written in following form by adding and subtracting $\hat{J}_p\dot{\theta}$ and using Property 3
\begin{equation}
\dot{e}_p  = -k_1 \phi(e_p) e_p - W_j \tilde{\zeta}_{j} + \hat{J}_p\eta
\label{eq:e_p_dot}
\end{equation}
where $\tilde{\zeta}_j = \zeta_j - \hat{\zeta}_j \in \mathbb{R}^{m_1}$ is the parameter estimation error. To compute the error dynamics of the auxiliary error $\eta$, taking the time derivative of (\ref{eq:eta}) yields, $\dot{\eta} =\frac{d}{dt}( \hat{J}_{p}^{-1}(
\dot{p}_{h} + k_1 \phi(e_p) e_p)) - \ddot{\theta}$.
Pre-multiplying $M(\theta)$ to $\dot{\eta}$ 
and substituting for $M\ddot{\theta}$ from (\ref{eq:robot_EL_eq}) results in
\begin{align}
M \dot{\eta} &= M \frac{d}{dt}( \hat{J}_{p}^{-1}(
\dot{p}_{h} + k_1 \phi e_p))+C(\theta,\dot{\theta})\dot{\theta}+f(\dot{\theta}) \nonumber \\
&+G(\theta) - \tau_r(t)  = -C(\theta,\dot{\theta})\eta - \tau_r + W_y\zeta_y  \label{eq:open_loop_eta}
\end{align}
where $W_y\zeta_y = M \frac{d}{dt}( \hat{J}_{p}^{-1}(
\dot{p}_{h} + k_1 \phi(e_p) e_p)) + C(\theta,\dot{\theta})\dot{\theta}+f(\dot{\theta}) +G(\theta) + C(\theta,\dot{\theta})\eta $, $W_y \in \mathbb{R}^{n\times m_2}$, $\zeta_y \in \mathbb{R}^{m_2}$ are parameters of the dynamics, and a corresponding parametrization using measurable signals is given by
\begin{align}
    W_{y_{T}}\zeta_{y} &= M \frac{d}{dt}( \hat{J}_{p}^{-1}(
\dot{p}_{h_{T}} + k_1 \phi(e_{p_{T}}) e_{p_{T}})) +C(\theta,\dot{\theta})\dot{\theta} \nonumber \\
&+f(\dot{\theta})+G(\theta) + C(\theta,\dot{\theta})\eta_T 
\end{align}
\subsection{Adaptive Synchronization Control}
Since only time-delayed human trajectory information is measurable, the synchronizing robot control torque is designed as
\begin{equation}
\tau_r(t) =W_{y_{T}}\hat{\zeta}_{y} + k_r\eta_T  +k_{\phi}\hat{J}_p^T\phi(e_{p_{T}}) e_{p_{T}}
\label{eq:robot_input}
\end{equation}
where $k_r$, $k_{\phi} \in \mathbb{R}^{+}$ are constant control gains, $\hat{\zeta}_y \in \mathbb{R}^{m_2}$ are parameter estimates. The update law for $\hat{\zeta}_y$ is designed as 
\begin{align}
\dot{\hat{\zeta}}_y = \Gamma_2 W_{y_{T}}^T \eta_T - \alpha_{s4}\Gamma_2 \hat{\zeta}_y
\label{eq:zeta_y_update}
\end{align}
where $\Gamma_2 \in \mathbb{R}^{m_2 \times m_2 }$, $\alpha_{s4} \in \mathbb{R}^{+}$ are the gains. The integral concurrent learning (ICL)-based parameter update law for $\hat{\zeta}_j$ to compute $\hat{J}_p$ is designed as 
\begin{align}
\dot{\hat{\zeta}}_j = -\Gamma_1 W_j^T \phi(e_{p_{T}}) e_{p_{T}} +\! k \Gamma_1 \sum_{i=1}^{N} [\mathcal{Y}_j^T(\mathcal{U}_j \!-\mathcal{Y}_j \hat{\zeta}_j)] \label{eq:ICLUpdateLaw}
\end{align}
where $\Gamma_1 \in \mathbb{R}^{m_1 \times m_1}$, $k \in \mathbb{R}^{+}$ are the gains, and
$\mathcal{U}_j = p(t)-p(t-\Delta t) = \int_{max(t -\Delta t , 0)}^{t} \dot{p} (\xi) d\xi$, 
$\mathcal{Y}_j = \int_{max(t -\Delta t , 0)}^{t}\! W_j(\theta, \dot{\theta})d\xi$ 
such that using (\ref{eq:der_position}) and (\ref{eq:kinematic_pramet}), following expression can be derived
\begin{equation}
\mathcal{U}_j = \mathcal{Y}_j \zeta_j \label{eq:UZeta}
\end{equation}
\begin{assumption} \label{assum:FiniteExcitation}
The system is sufficiently excited over a finite time period $T_{fe}$. This implies there exists $ \lambda_{m} > 0$ and $T_{fe} > \Delta t : \forall \Delta t \geq T_{fe}$, $\lambda_{\mathrm{min}} \left\{ \sum_{i=1}^N \mathcal{Y}_i^T \mathcal{Y}_i\right\} \geq \lambda_m.$
\end{assumption}

\subsection{Closed-loop Error Dynamics}
Even though the controller uses time-delayed information, the stability and convergence analysis of the synchronization error with respect to current time is analyzed. To facilitate the stability analysis, the torque is written in terms of $e_p(t)$ and $\eta(t)$
by adding and subtracting $k_{\phi}\hat{J}_p^T\phi(e_{p}) e_{p}$ 
\begin{align}
&\tau_r(t) \!=\! W_{y_{T}}\hat{\zeta}_{y} + k_r\eta + k_r\hat{J}_{p}^{-1} 
(\dot{p}_{h_{T}} + k_1 \phi(e_{p_{T}}) e_{p_{T}} -
\dot{p}_{h} \nonumber \\ 
&-k_1\phi(e_p) e_p)\!+\!k_{\phi}\hat{J}_p^T\phi(e_{p})e_{p}\!-\!k_{\phi}\hat{J}_p^T(\phi(e_{p})e_{p}\!-\!\phi(e_{p_{T}}) e_{p_{T}})
\label{eq:robot_input2}
\end{align}
After substituting (\ref{eq:robot_input2}) into (\ref{eq:open_loop_eta}) and addition and subtraction of $W_{y_{T}}\zeta_y$, the following closed-loop dynamics for $\dot{\eta}$ is obtained
\begin{align}
M \dot{\eta} &= -C(\theta,\dot{\theta})\eta + W_{y_{T}}\tilde{\zeta}_y + (W_y - W_{y_{T}})\zeta_y -[ k_r\eta \nonumber \\
&+ k_r 
\hat{J}_{p}^{-1}(\dot{p}_{h_{T}} + k_1 \phi(e_{p_{T}}) e_{p_{T}} -
\dot{p}_{h} - k_1 \phi(e_p) e_p)    \nonumber \\ 
&+ k_{\phi}\hat{J}_p^T\phi(e_{p}) e_{p}- k_{\phi}\hat{J}_p^T(\phi(e_{p}) e_{p} - \phi(e_{p_{T}}) e_{p_{T}})]
\label{eq:eta_closed_loop_alt}
\end{align}
where $\tilde{\zeta}_y(t) = \zeta_y  - \hat{\zeta}_y(t) \in \mathbb{R}^{m_2}$. Using (\ref{eq:zeta_y_update}), (\ref{eq:ICLUpdateLaw}), and (\ref{eq:UZeta}), the parameter estimation error dynamics can be written as
\begin{equation}
    \dot{\tilde{\zeta}}_y = -  \Gamma_2 W_{y_{T}}^T \eta_T + \alpha_{s4}\Gamma_2 \hat{\zeta}_y
    \label{eq:zeta_y_ErrorDynamics}
\end{equation}
\begin{equation}
    \dot{\tilde{\zeta}}_j =  
    \Gamma_1 W_j^T  \phi_T e_{p_{T}} -\! k \Gamma_1 \sum_{i=1}^{N} [\mathcal{Y}_j^T\mathcal{Y}_j]\tilde{\zeta}_j \label{eq:zetaErrorDynamics}
\end{equation}
\subsection{Stability Analysis}
\label{sec: sym_StabilityAnalysis}
In this section, stability analysis of the safe synchronization controller is presented for the error at the current time. Let's define an auxiliary vector $z(t) = [\sqrt{V_1},\eta^T \;\tilde{\zeta}_{y}^T \;\tilde{\zeta}_{j}^T \;  \sqrt{P}_1, \sqrt{P}_2, \sqrt{P}_3]^T \in \mathcal{D} \subset \mathbb{R}^{n+m_1+m_2+4}$, where $V_1$ is defined in the next section and the LK functionals $P_i \in \mathbb{R}^{n}$ are defined for delay systems analysis as follows
\begin{align}
P_1 &= \frac{K_{LK_{1}} \omega_1}{2}\int_{t_T}^{t}\int_{s}^{t} \Vert \frac{d}{dl}\dot{p}_h(l)\Vert^2 dlds
\label{eq:P_1}  \\
P_2 &= \frac{K_{LK_{2}} \omega_2}{2} \int_{t_T}^{t}\int_{s}^{t}\Vert \frac{d}{dl}(\phi(e_p)(l) e_p(l))\Vert^2 dlds 
\label{eq:P_2} \\
P_3  &= \frac{K_{LK_{3}} \omega_3}{2} \int_{t_T}^{t}\int_{s}^{t} \Vert \dot{p}_h(l) \Vert^2 dl ds \label{eq:P_3}
\end{align}
and $\omega_i \in \mathbb{R}^+$ and $K_{LK_{i}}\in \mathbb{R}^+$ are constants.

\begin{theorem}
For the system defined in (\ref{eq:robot_EL_eq}), the synchronization controller in (\ref{eq:robot_input}) and adaptive laws in (\ref{eq:zeta_y_update})-(\ref{eq:ICLUpdateLaw}) ensure all the closed-loop signals are bounded and the end-effector states SGUUB synchronize in the sense 
\begin{equation}
    \Vert e_p \Vert \leq \varepsilon_3 \Vert e_p(0) \Vert e^{-\varepsilon_2t} + \varepsilon_1 \label{eq:syncBound}
\end{equation} 
where $\varepsilon_1$, $\varepsilon_2$ and $\varepsilon_3$ are positive constants, and $e_p(t) \in \mathcal{Z}$, if $e_p(0) \in \mathcal{Z}$, where $\mathcal{Z}$ is subsequently defined set, provided Assumptions \ref{assum:Assumption1}-\ref{assum:FiniteExcitation} and following gain conditions are satisfied 
\begin{align}
\alpha_{s4} -\frac{\alpha_{s4}}{4\gamma_\zeta}- \frac{\gamma_5^2}{4} >0, \frac{K_{LK_{3}}}{2} - c_{LK_{3}}>0, \omega_i >T \label{eq:suffCond}
\end{align}
where $\gamma_5$, $\gamma_\zeta$, and $c_{LK_{3}}$ are subsequently defined constants.
\end{theorem}
\begin{proof}
Consider a positive definite BLF defined by
\begin{equation}
    V_1(e_p) = \sum_{i=1}^n V_i =\frac{1}{2} \Sigma_{i=1}^{n} \mathrm{log} k_{m_i}^2 \phi_{i}, \quad \forall i \in \{1,..,n\} \label{eq:BLF}
\end{equation}
such that $\mathrm{lim}_{|e_{pi}| \rightarrow k_{mi}}V_1(e_{pi}) = \infty$. Consider a combined Barrier Lyapunov-Krasovskii functional candidate $V(z,t) = V_1(e_p) + V_2$ defined as $V(z,t) : \mathcal{D} \times [0,\infty) \rightarrow \mathbb{R}^{+}$, which is a positive definite Lyapunov functional, continuously differentiable in $z$, locally Lipschitz in $t$, given by
\begin{align}
V(z,t) &= V_1 + \frac{1}{2} \eta^TM(\theta)\eta + \frac{1}{2} \tilde{\zeta}_j^T \Gamma_1^{-1}\tilde{\zeta}_j   \nonumber \\ 
&+ \frac{1}{2} \tilde{\zeta}_y^T \Gamma_2^{-1}\tilde{\zeta}_y + P_1 + P_2 + P_3 \label{eq:LyapunovFn}
\end{align}
Taking the time derivative of $V(z,t)$ results in
\begin{align}
\dot{V}(z) &=  \Sigma_{i=1}^{3} e_{p_{i}} \phi_{i}\dot{e}_{p_{i}} + \frac{1}{2} \eta^T\dot{M}(\theta)\eta + \eta^TM(\theta)\dot{\eta} \nonumber \\ 
&+ \tilde{\zeta}_j^T\Gamma_1^{-1}\dot{\tilde{\zeta}}_j  + \tilde{\zeta}_y^T \Gamma_2^{-1}\dot{\tilde{\zeta}}_y   + \dot{P}_1  + \dot{P}_2  + \dot{P}_3
\end{align}
Substituting from (\ref{eq:e_p_dot}) and (\ref{eq:eta_closed_loop_alt}),
\begin{align}
&\dot{V}(z) =  e_p^T \phi (-k_1 \phi(e_p) e_p - W_j \tilde{\zeta}_{j} + k_{\phi} \hat{J}_p\eta) \nonumber \\ 
&+\eta^T( W_{y_{T}}\tilde{\zeta}_y + (W_y - W_{y_{T}})\zeta_y - k_r\eta - k_r 
\hat{J}_{p}^{-1}(\dot{p}_{h_{T}} \nonumber \\ 
&+ k_1 \phi(e_{p_{T}}) e_{p_{T}} -
\dot{p}_{h} - k_1 \phi(e_p) e_p) -  k_{\phi}\hat{J}_p^T\phi(e_{p}) e_{p}  \nonumber \\ 
&+ k_{\phi}\hat{J}_p^T(\phi(e_{p}) e_{p} - \phi(e_{p_{T}}) e_{p_{T}})) + \tilde{\zeta}_j^T\Gamma_1^{-1}\dot{\tilde{\zeta}}_j  \nonumber \\ 
& + \tilde{\zeta}_y^T \Gamma_2^{-1}\dot{\tilde{\zeta}}_y + \dot{P}_1  + \dot{P}_2 +\dot{P}_3
\end{align}
Using the derivatives of $P_1$, $P_2$ and $P_3$, (\ref{eq:zeta_y_ErrorDynamics}) and (\ref{eq:zetaErrorDynamics}), and 
$\dot{p}_{h} - \dot{p}_{hT} = \int_{t_T}^t\ddot{p}_h(l)dl, 
    \phi(e_{p})e_{p} - \phi(e_{pT})e_{pT} = \int_{t_T}^t \frac{d}{dl}\phi(e_p(l))e_p(l)dl, $ 
$\dot{V}(z)$ can be written as 
\begin{align}
&\dot{V}(z) =  -k_1 \sum_{i=1}^n\phi_i^2(e_p) e_{pi}^2 - e_p^T \phi W_j \tilde{\zeta}_{j} \nonumber \\ 
&+ \eta^T[ -(W_{y_{T}}- W_y)\zeta_{y} - k_r\eta + W_{y_{T}}\tilde{\zeta}_y] + \eta^T [k_r\hat{J}_{p}^{-1} 
\nonumber \\ 
&\times\int_{t_T}^{t} \ddot{p}_h(l) dl \!+\! (k_r k_1 \hat{J}_{p}^{-1} \!+\! k_{\phi}\hat{J}_p^T)\int_{t_T}^{t}\frac{d}{dt}(\phi(e_p)(l) e_p(l)) dl] \nonumber \\
&+ \tilde{\zeta}_j^T[ W_j^T  \phi e_{p} + W_j^T  (\phi_T e_{p_{T}} -  \phi e_{p}) - k  \sum_{i=1}^{N} [\mathcal{Y}_j^T\mathcal{Y}_j]\tilde{\zeta}_j]  \nonumber \\ 
&+ \tilde{\zeta}_y^T[- W_{y_{T}}^T \eta + W_{y_{T}}^T (\eta - \eta_T) + \alpha_{s4} \hat{\zeta}_y]  \nonumber \\ 
&  + \frac{K_{LK_{1}}\omega_1T}{2}\Vert \ddot{p}_h(t) \Vert^2-\frac{K_{LK_{1}}\omega_1}{2}\int_{t_T}^{t} \Vert\ddot{p}_h(l)\Vert^2 dl\nonumber \nonumber \\ 
&+\frac{K_{LK_{2}}\omega_2T}{2}\Vert\dot{\phi}(e_p)e_p(t) + \phi(e_p)(t)\dot{e}_p(t)\Vert^2\nonumber \\ 
&- \frac{K_{LK_{2}}\omega_2}{2} \int_{t_T}^{t}\Vert\frac{d}{dt}(\phi(e_p)(l) e_p(l))\Vert^2  dl \nonumber \\
&+ \frac{K_{LK_{3}}\omega_3T}{2}\Vert \dot{p}_h(t) \Vert^2-\frac{K_{LK_{3}}\omega_3}{2}\int_{t_T}^{t} \Vert\dot{p}_h(l)\Vert^2 dl \label{eq:VdotInter}
\end{align}
Canceling out cross terms, using $\Vert\sum_{j=1}^N\mathcal{Y}_j^T\mathcal{Y}_j\Vert \geq\lambda_m$, $k_1 = k_{1_{V_{1}}} + k_{1_{e_{p}}}$, $2V_1 < \Sigma_{i=1}^{n} \phi_ie_{p_{i}}^2$, and the Appendices provided in the supplemental document \cite{BhattacharyaLCSSSuppl} developed using \cite{de1997adaptive}, results in \vspace{-6pt}
\begin{align}
&\dot{V}(z) \leq  -2k_{1_{V_{1}}}\sum_{i=1}^n\phi_iV_i - k_{1_{e_{p}}}\Vert\phi^2\Vert\Vert e_p\Vert^2 - k_r\Vert \eta \Vert^2 \nonumber \\ 
&-k\lambda_m\Vert\tilde{\zeta}_j\Vert^2-\alpha_{s4}\Vert \tilde{\zeta}_y \Vert^2 + \alpha_{s4}\tilde{\zeta}_y^T\zeta_y+ \eta^T [k_r\hat{J}_{p}^{-1}\int_{t_T}^{t} \ddot{p}_h(l) dl 
\nonumber \\ 
&+ (k_r k_1 \hat{J}_{p}^{-1} + k_{\phi}\hat{J}_p^T)\int_{t_T}^{t}\frac{d}{dt}(\phi(e_p)(l) e_p(l)) dl] \nonumber \\
&+\bar{m}\bar{j}_{p_{inv}}\bar{p}_3 
\Vert \eta \Vert \!+\! \rho_9(\Vert z_1\Vert, \Vert \dot{\theta}\Vert)\Vert \eta \Vert^2 \!+\! c_{LK_{1}}  
\Vert \int_{t_T}^{t} \ddot{p}_h(l) dl \Vert ^2\nonumber \\
&+ c_{LK_{2}} \Vert\int_{t_T}^{t} \frac{d}{dl} \phi(e_p)(l) e_p(l) dl\Vert^2 +c_{LK_{3}} \Vert \int_{t_{t}}^t \dot{p}_h dl \Vert^2\nonumber \\
& +(\frac{\bar{p}_1 w_j }{\gamma_{11}^2}+\frac{\bar{p}_1 w_j }{\gamma_{13}^2}) \Vert \tilde{\zeta}_{j} \Vert^2 +\frac{\gamma_6^2}{4}w_j^2\Vert \dot{\theta} \Vert^2 \Vert\tilde{\zeta}_j \Vert^2+\frac{\gamma_5^2}{4}\Vert\tilde{\zeta}_y\Vert^2\nonumber \\
& + \frac{1}{\gamma_6^2}\Vert\int \frac{d}{dt} \phi(e_p)e_p dl \Vert^2
+ \frac{2j_{p_{i}}^2}{\gamma_5^2}\rho_2^2(\Vert y_T\Vert) 
\Vert \int_{t_T}^{t} \ddot{p}_h(l) dl \Vert ^2  \nonumber \\ 
&+ \frac{ 2k_1^2 j_{p_{i}}^2}{\gamma_5^2}\rho_2^2(\Vert y_T\Vert)\Vert \int_{t_T}^{t}\frac{d}{dt}(\phi(e_p)(l) e_p(l)) dl\Vert ^2-\frac{K_{LK_{1}}\omega_1}{2} \nonumber \\
& \times\int_{t_T}^{t} \Vert\ddot{p}_h(l)\Vert^2 dl- \frac{K_{LK_{2}}\omega_2}{2} \int_{t_T}^{t}\Vert\frac{d}{dt}(\phi(e_p)(l) e_p(l))\Vert^2   dl \nonumber \\ 
& -\frac{K_{LK_{3}}\omega_3}{2}\int_{t_T}^{t} \Vert\dot{p}_h(l)\Vert^2 dl + Tc + T\rho_1(\Vert y \Vert)\Vert y \Vert^2
\end{align}
where  $c=c_1+c_2$, $c_1=\frac{K_{LK_{1}}\omega_1T}{2}p_1^2$, $c_2= \frac{K_{LK_{3}}\omega_3T}{2}p_2^2$. Using Cauchy-Schwarz and Young's inequalities, $\Vert \phi \Vert \geq \frac{1}{\mathrm{max}\{k_{m_{i}}^2\}} = \bar{\phi}$, and bounds on $\hat{J}_p$ and $\hat{J}_p^{-1}$ from Assumption \ref{assum:Assumption3}, the following upper bound can be developed
\begin{align}
&\dot{V}(z) \leq -k_{1_{e_{p}}}\bar{\phi}^2\Vert e_p\Vert^2 - k_r\Vert \eta \Vert^2 -k\lambda_m\Vert\tilde{\zeta}_j\Vert^2 -\alpha_{s4}\Vert \tilde{\zeta}_y \Vert^2\nonumber \\ 
& + \alpha_{s4}\tilde{\zeta}_y^T\zeta_y +\frac{\gamma_5^2}{4}\Vert\tilde{\zeta}_y\Vert^2  + T\rho_1(\Vert y \Vert)\Vert e_p \Vert^2 + \xi_1(\Vert y \Vert)\Vert \eta \Vert^2 \nonumber \\
& +\bar{m}\bar{j}_{p_{inv}}\bar{p}_3 
\Vert \eta \Vert + \xi_2(\Vert y \Vert) \Vert\tilde{\zeta}_{j} \Vert^2 + \xi_3(\Vert y_T \Vert) \Vert\int_{t_T}^{t} \ddot{p}_h(l) dl \Vert^2 \nonumber \\ 
&+ \xi_4(\Vert y_T \Vert) \Vert\int_{t_T}^{t} \frac{d}{dt}(\phi(e_p)(l) e_p(l)) dl \Vert^2 +c_{LK_{3}} \Vert \int_{t_{t}}^t \dot{p}_h dl \Vert^2  \nonumber \\ 
&  -\frac{K_{LK_{1}}\omega_1}{2}\int_{t_T}^{t} \Vert \ddot{p}_h \Vert^2 dl - \frac{K_{LK_{2}}\omega_2}{2} \int_{t_T}^{t} \Vert \frac{d}{dt}(\phi(e_p) e_p \Vert^2 dl \nonumber \\
&-\frac{K_{LK_{3}}\omega_3}{2}\int_{t_T}^{t} \Vert \dot{p}_h \Vert^2 dl - 2k_{1_{V_{1}}}\bar{\phi}V_1 + Tc 
\end{align}
where $y(t)=[e_p^T\;\eta^T\;\tilde{\zeta}_j^T]^T$, $z_1(t)= [e_p^T\;\eta^T]^T$, and using (\ref{eq:e_p_dot}), the functions $\xi_i$ are defined as
\begin{align}
    \xi_1(\Vert y \Vert) &= \frac{(\gamma_1^2+2\gamma_2^2)}{4} + \rho_9(\Vert z_1\Vert, \Vert \dot{\theta}\Vert) + T\rho_1(\Vert y \Vert)  \nonumber \\
    \xi_2(\Vert y \Vert) &= \frac{\bar{p}_1 w_j}{\gamma_{11}^2}+\frac{\bar{p}_1 w_j }{\gamma_{13}^2} + \frac{\gamma_6^2}{4}w_j^2\Vert \dot{\theta} \Vert^2 + T\rho_1(\Vert y \Vert) \nonumber
\end{align}
\begin{align}
    &\xi_3(\Vert y_T \Vert) \!=\!\frac{k_r^2 j_{p_{i}}^2}{\gamma_1^2}+ c_{LK_{1}} + \frac{2j_{p_{i}}^2}{\gamma_5^2}\rho_2^2(\Vert y_T\Vert) \nonumber \\
    &\xi_4(\Vert y_T \Vert) \!=\!\frac{(k_r^2k_1^2 +k_{\phi}^2)j_p^2}{\gamma_2^2} + c_{LK_{2}} + \frac{1}{\gamma_6^2} + \frac{ 2k_1^2 j_{p_{i}}^2}{\gamma_5^2}\rho_2^2(\Vert y_T\Vert) \nonumber
\end{align}
Writing $k_r = k_{r1} + k_{r2}$, completing the squares on $\eta$, and using Young's inequality for $\tilde{\zeta}_y$ term yields
\begin{align}
&\dot{V}(z) \leq  -2k_{1_{V_{1}}}\bar{\phi}V_1-k_{1_{e_{p}}}\bar{\phi}^2\Vert e_p\Vert^2 - k_{r1}\Vert \eta \Vert^2 -k\lambda_m\Vert\tilde{\zeta}_j\Vert^2 \nonumber \\ 
& -\left(\alpha_{s4} -\frac{\alpha_{s4}}{4\gamma_\zeta}- \frac{\gamma_5^2}{4}\right)\Vert \tilde{\zeta}_y \Vert^2  + T\rho_1(\Vert y \Vert)\Vert e_p \Vert^2 \nonumber \\
&+ \xi_1(\Vert y \Vert)\Vert \eta \Vert^2  + \xi_2(\Vert y \Vert) \Vert\tilde{\zeta}_{j} \Vert^2 + \xi_3(\Vert y_T \Vert) \Vert\int_{t_T}^{t} \ddot{p}_h(l) dl \Vert^2 \nonumber \\ 
&+ \xi_4(\Vert y_T \Vert) \Vert\int_{t_T}^{t} \frac{d}{dt}(\phi(e_p)(l) e_p(l)) dl \Vert^2 +c_{LK_{3}} \Vert \int_{t_{t}}^t \dot{p}_h dl \Vert^2  \nonumber \\ 
&  -\frac{K_{LK_{1}}\omega_1}{2}\int_{t_T}^{t} \Vert \ddot{p}_h \Vert^2 dl - \frac{K_{LK_{2}}\omega_2}{2} \int_{t_T}^{t} \Vert \frac{d}{dt}(\phi(e_p) e_p \Vert^2 dl \nonumber \\
&-\frac{K_{LK_{3}}\omega_3}{2}\int_{t_T}^{t} \Vert \dot{p}_h \Vert^2 dl + \epsilon_1
\end{align}
where $\gamma_{\zeta} \in \mathbb{R}^+$ and $\epsilon_1 =Tc +\frac{(\bar{m}\bar{j}_{p_{inv}}\bar{p}_3)^2}{4k_{r2}} + \alpha_{s4}\gamma_{\zeta}\Vert\zeta_y\Vert^2$ are constants. 
Using $\Vert \int_{t_T}^t \ddot{p}_h(l)dl\Vert^2 \leq T \int_{t_T}^t \Vert \ddot{p}_h(l)\Vert^2dl$, $\Vert \int_{t_T}^t \frac{d}{dt}\phi e_p dl\Vert^2 \leq T \int_{t_T}^t \Vert \frac{d}{dt} \phi e_p \Vert^2dl$, $\Vert \int_{t_T}^t \dot{p}_h(l)dl\Vert^2 \leq T \int_{t_T}^t \Vert \dot{p}_h(l)\Vert^2dl$, $\dot{V}(z)$ can be upper bounded as
\begin{align}
&\dot{V}(z) \leq  -\left(k_m -T\rho_1(\Vert y \Vert) \right)\Vert e_p\Vert^2 - \left(k_m - \xi_1(\Vert y \Vert)\right)\Vert \eta \Vert^2 \nonumber \\
&-(k_m - \xi_2(\Vert y \Vert))\Vert\tilde{\zeta}_j\Vert^2 -\left(\alpha_{s4} -\frac{\alpha_{s4}}{4\gamma_\zeta}- \frac{\gamma_5^2}{4}\right)\Vert \tilde{\zeta}_y \Vert^2 \nonumber \\
& - \left(\frac{K_{LK1}}{2}-\xi_3(\Vert y_T \Vert) \right)T \int_{t_T}^{t} \Vert\ddot{p}_h(l)\Vert^2 dl \nonumber \\ 
&- \left(\frac{K_{LK2}}{2}-\xi_4(\Vert y_T \Vert)\right) T \int_{t_T}^{t} \Vert \frac{d}{dt}(\phi(e_p)(l) e_p(l)) \Vert^2 dl   \nonumber \\
&- \left(\frac{K_{LK_{3}}}{2} - c_{LK_{3}}\right) T \int_{t_{t}}^t \Vert\dot{p}_h \Vert^2 dl -2k_{1_{V_{1}}}\bar{\phi}V_1 + \epsilon_1
\end{align}
where $k_m =\mathrm{min}\{k_{1_{e_{p}}}\bar{\phi}^2,k_{r1},k\lambda_m\}$, $-\omega_1 < -T$, $-\omega_2 < -T$ and $-\omega_3 < -T$. Utilizing following inequalities
\begin{align}
&-\int_{t_T}^{t} \int_{s}^{t}\Vert \frac{d}{dl}\dot{p}_h(l)\Vert^2 dl ds \geq - T\int_{t_T}^{t}  \Vert \ddot{p}_h(l) \Vert^2 dl  \nonumber
\\
&-\int_{t_T}^{t}\!\int_{s}^{t} \Vert \frac{d}{dl}(\phi(e_p)e_p)) \Vert^2 \!dl ds \!\geq\!  -T \int_{t_T}^{t}  \Vert \frac{d}{dl}(\phi(e_p)e_p) \Vert^2 \!dl   \nonumber 
\\
&-\int_{t_T}^{t} \int_{s}^{t} \Vert \phi(e_p)e_p) \Vert^2 dl ds \geq  -T \int_{t_T}^{t} \Vert\phi(e_p)e_p)\Vert^2 dl  \nonumber 
\end{align}
and using (\ref{eq:P_1}), (\ref{eq:P_2}), and (\ref{eq:P_3}) yields
\begin{align}
&\dot{V}(z) \leq  -\left(k_m -T\rho_1(\Vert y \Vert) \right)\Vert e_p\Vert^2 - \left(k_m - \xi_1(\Vert y \Vert)\right)\Vert \eta \Vert^2 \nonumber \\
&-(k_m - \xi_2(\Vert y \Vert))\Vert\tilde{\zeta}_j\Vert^2 -\left(\alpha_{s4} -\frac{\alpha_{s4}}{4\gamma_\zeta}- \frac{\gamma_5^2}{4}\right)\Vert \tilde{\zeta}_y \Vert^2 \nonumber \\
& - \left(\frac{K_{LK1}}{2}-\xi_3(\Vert y_T \Vert) \right) P_1 - \left(\frac{K_{LK2}}{2}-\xi_4(\Vert y_T \Vert)\right)  P_2  \nonumber \\
&- \left(\frac{K_{LK_{3}}}{2} - c_{LK_{3}}\right) P_3  -2k_{1_{V_{1}}}\bar{\phi}V_1 + \epsilon_1
\end{align}
From the definition of $z$, following bound on $\dot{V}$ can be written
\begin{align}
\dot{V}\leq -\beta_1V +\epsilon_1
\end{align}
where $\beta_1= \mathrm{min}\left(2k_{1_{V_{1}}}\bar{\phi}, 2\frac{k_m - \xi_1(\Vert y \Vert)}{\bar{m}}\right.$, $2\frac{k_m - \xi_2(\Vert y \Vert)}{\lambda_{\mathrm{max}}(\Gamma_1^{-1})}, 2\frac{\alpha_{s4} -\frac{\alpha_{s4}}{4\gamma_\zeta}- \frac{\gamma_5^2}{4}}{\lambda_{\mathrm{max}}(\Gamma_2^{-1})}, \frac{K_{LK1}}{2}-\xi_3(\Vert y_T \Vert)$, $\left. \frac{K_{LK2}}{2}-\xi_4(\Vert y_T \Vert), \frac{K_{LK_{3}}}{2} - c_{LK_{3}}\right)$.
Consider the sets 
\begin{align}
&\mathcal{S}_y \!= \! \{y \in \mathbb{R}^{2n+m_1} | \Vert y \Vert \!< \!\mathrm{min}\left(\rho_1^{-1}(k_m),\xi_1^{-1}(k_m), \xi_2^{-1}(k_m) \right) \} \nonumber \\
&\mathcal{S}_{yT} =  \{ y_T \in \mathbb{R}^{2n+m_1} |   \nonumber \\
&\Vert y_T\Vert < \mathrm{min}\left(\xi_3^{-1}\left(\frac{K_{LK1}}{2}\right ), \xi_4^{-1}\left(\frac{K_{LK2}}{2}\right) \right) \}
\end{align}
Within the sets $\mathcal{S} \cup \mathcal{S}_T$, $\beta_1 \geq \bar{\beta}_1$, where $\bar{\beta}_1$ is a positive constant and following bound can be developed
$V(z,t) \leq  V(0) e^{-\bar{\beta}_1t}+\frac{\epsilon_1}{\bar{\beta}_1}[1-e^{-\bar{\beta}_1t}]$.
If $y(0) \in \mathcal{S}$ and $y_T(0) \in \mathcal{S}_T$, then the gains can be selected according to the sufficient conditions in (\ref{eq:suffCond}) to obtain the bound in (\ref{eq:syncBound}) yielding a SGUUB stability. Using standard signal chasing argument, all the closed-loop signals and torque input are bounded. We have $V_1(0) \geq V_1(t) = \frac{1}{2}\Sigma_{i=1}^{n}\mathrm{log}\frac{k_{m_{i}}^2}{k_{m_{i}}^2-e_{p_{i}}^2}$, 
Since $V_1(t)\leq  V(t)$, $\frac{1}{2}\Sigma_{i=1}^{n}\mathrm{log}\frac{k_{m_{i}}^2}{k_{m_{i}}^2-e_{p_{i}}^2} \leq  V(0) e^{-\bar{\beta}_1t}+\frac{\epsilon_1}{\bar{\beta}_1}[1-e^{-\bar{\beta}_1t}]$. 
After some algebraic manipulations, $\vert e_{p_{i}}(t)\vert \leq k_{m_{i}}\sqrt{1- e^{- \varrho}}$ such that
\begin{equation}
    \mathcal{Z} = \{  e_{p} \in \mathbb{R}^3: \vert e_{p_{i}} \vert \leq k_{m_{i}}\sqrt{1- e^{-\varrho}},\; i \in \{1,..,n\}
\end{equation} where $\varrho = -2( V(0) e^{-\bar{\beta}_1t}+\frac{\epsilon_1}{\bar{\beta}_1}[1-e^{-\bar{\beta}_1t}])$.
\end{proof}

\section{Simulations\label{sec:simulation}}
\begin{figure}
\begin{centering}
\makebox[1\linewidth][c]{\subfigure[]{\label{fig:error}\includegraphics[width=0.55\columnwidth]{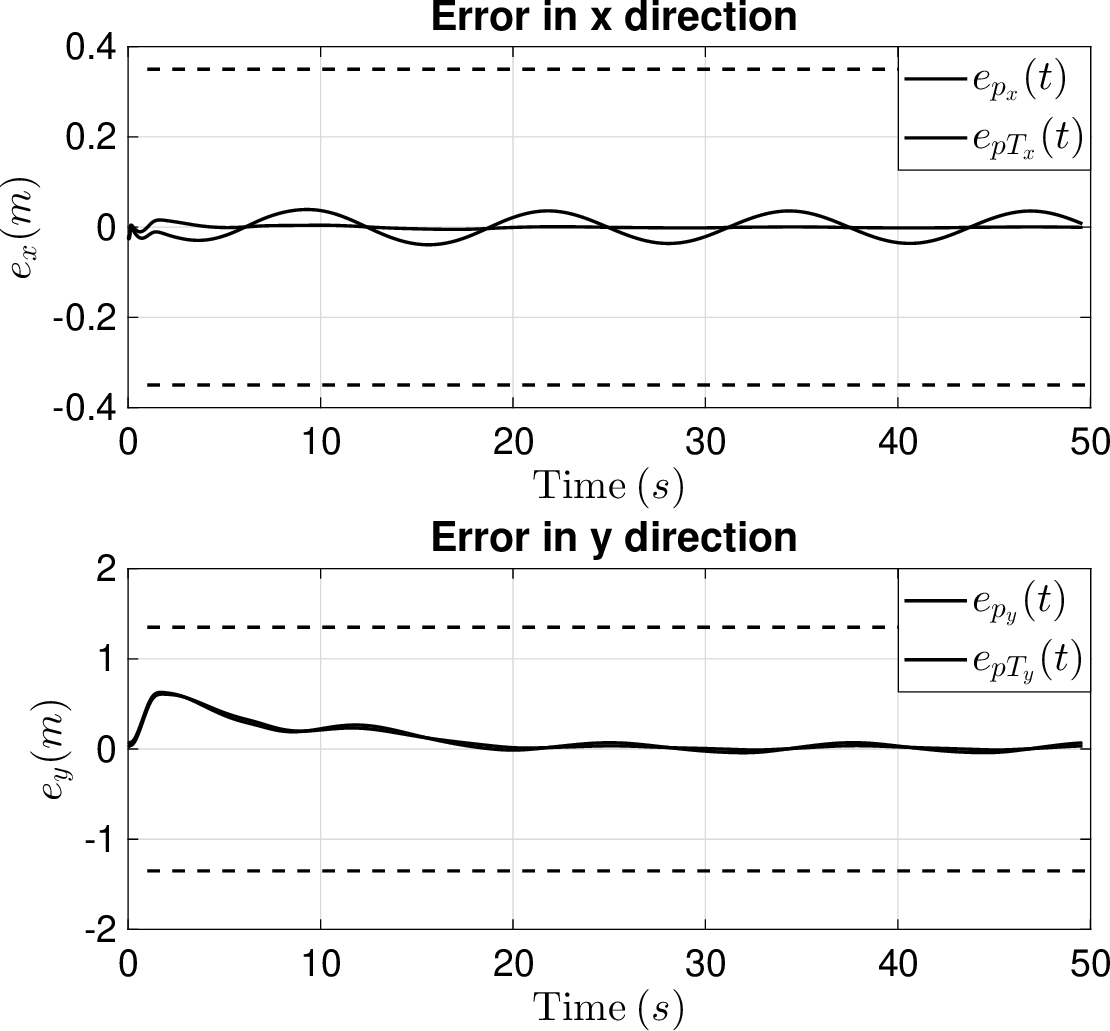}}\hspace{-1.0em}\subfigure[]{\label{fig:norms}\includegraphics[width=0.55\columnwidth]{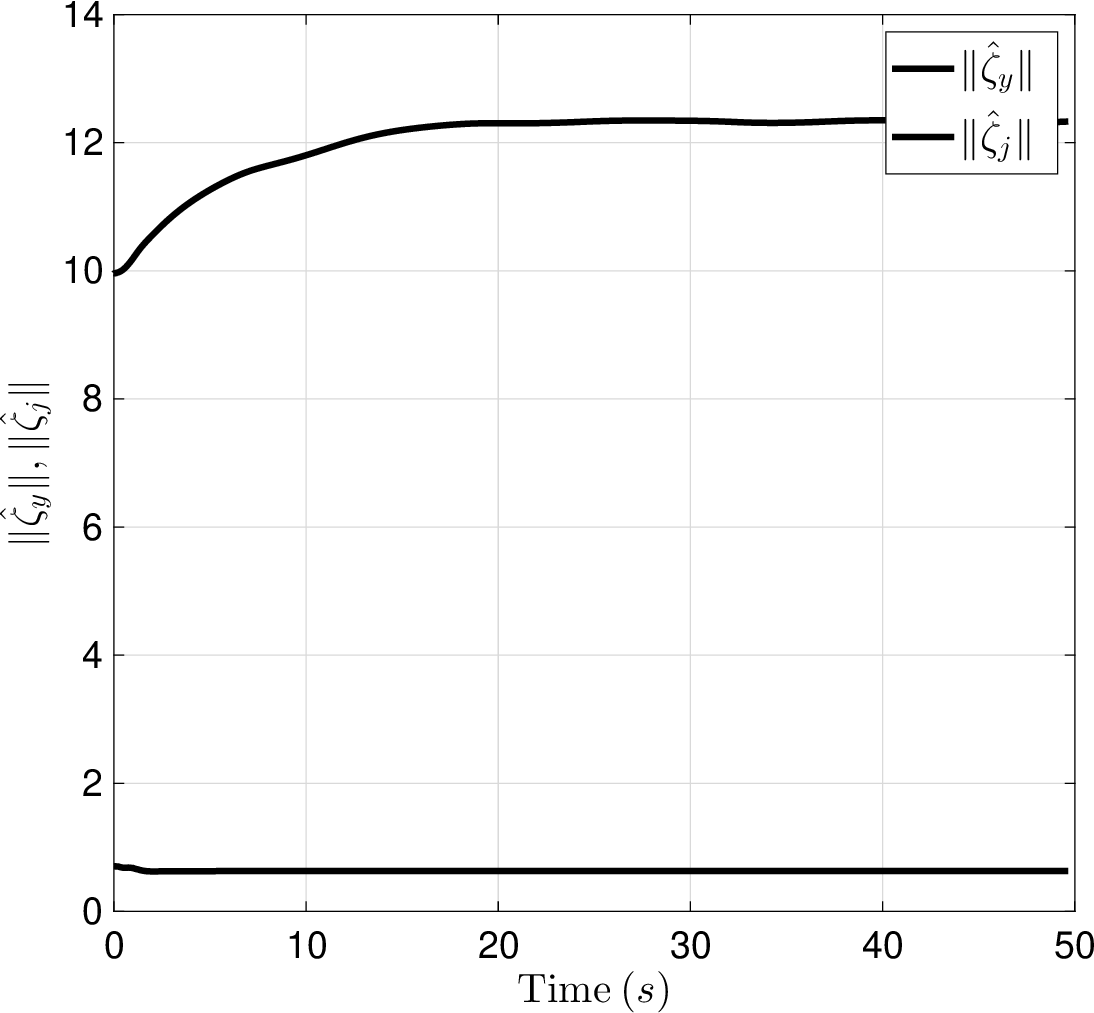}}}
\end{centering}
\centering{}\vspace{-8pt}
\caption{Results of simulation: (a) synchronization errors with respect to human trajectory at current time and at delayed time, (b) parameter estimate norms.\label{Simulation1}}
\end{figure}

The adaptive synchronization controller developed in task space is tested in simulation using a 2-DOF robot model, coded in MATLAB R2025a. The robot is modeled with link masses $0.558$kg and $0.291$kg and their lengths $0.85$m and $1.3$m, respectively, considering effects due to gravity $g=9.81$m/sec$^2$ and friction. The simulation is performed with symmetric constraints on the Cartesian coordinates of the robot's end-effector for a duration of $50$sec at a frequency of $100$Hz. In order to address the delay in human trajectory estimation, a time delay of $0.45$sec is introduced. The synchronization trajectory obtained from the human intent estimation model is described as $p_h(t) = [0.55 + 0.2\cos(0.5t), 0.25 + 0.2\sin(0.5t)]^{T}$. The controller gain parameters are selected as $k_r=0.1$, $k_{\phi}=1$, $k_{1}=0.8$ and $k=100$, $\Gamma_1=\mathbb{I}_{2\times2}$, $\Gamma_2=0.5\mathbb{I}_{7\times7}$, and $\alpha_{s4}=0.002$. The ICL term parameters are selected as $N=25$ and $\Delta t=0.2 \:\mathrm{sec}$, where $N$ is selected using $N\geq \lceil \frac{n.m_1}{n} \rceil$,  $\lceil \cdot \rceil$ denotes the ceiling function. The initial values of the parameters are set as $p(0)=[0.78 , 0.23]^T$, $\hat{\zeta}_j(0)=[0.5, 0.25]^T$ and $\hat{\zeta}_y(0)=[0.502, 0.082, 0.158, 0.0185, 9.5, 2.78, 0.0185]^T$. The bounds on the human trajectory are $|p_{h1}| < 0.75$ and $|p_{h2}| < 0.45$, and the robot end-effector limits are $|p_1| < 1.15$,  $|p_2| <1.85$, using which $k_{m_{1}}=0.4$ and  $k_{m_{2}}=1.4$. The robot joint angles are initialized within the constrained bounds. From the simulation results shown in Figs. \ref{fig:error}-\ref{fig:norms}, it is observed that the synchronization errors with respect to current human trajectory is ultimately bounded using the time delayed information and robot end-effector states remain within a prescribed bounds, $(-k_{m_{i}}, k_{m_{i}})$, $\forall i \in \{1,2\} $. From Fig. \ref{fig:error}, it is also observed that the error with respect to delayed time trajectory, $e_{pT}(t)$ converges to zero.
\vspace{-6pt}
\section{Conclusion\label{sec:Conclusion}}
In this paper, an adaptive trajectory synchronization controller is developed in task space to synchronize the robot's end-effector motion with the human hand motion in the presence of kinematic and dynamic uncertainty. Using only the available time-delayed information of the human hand trajectory, caused due to image processing or network communication delays, the controller synchronizes the robot trajectories in task space with the human trajectory within a bounded error, while ensuring that the robot trajectories remain within a prescribed safe set, which is established through Lyapunov stability analysis using Barrier Lyapunov function and LK functionals. Simulation studies demonstrate that the developed controller achieves the UUB bound, with kinematic parameter estimates converging to their true values.
\vspace{-7pt}
\appendices
\section{Development of a bounds on $\dot{P}_1$ , $\dot{P}_2$ and $\dot{P}_3$} \label{app:Appendix1}
The time derivatives of the LK functional  terms using Leibniz rule yield
\begin{equation}
\dot{P}_1 \leq Tc_1-\frac{K_{LK_{1}}\omega_1}{2}\int_{t_T}^{t} \ddot{p}_h^T(l)\ddot{p}_h(l) dl\nonumber
\end{equation}
where $c_1 \geq \frac{K_{LK_{1}}\omega_1}{2}\Vert \ddot{p}_h(t) \Vert^2 \in \mathbb{R}^+$ is a constant. 
Similarly, the derivative of $P_2$ can be computed as follows
\begin{align}
\dot{P}_2&=\frac{K_{LK_{2}}\omega_2T}{2}\Vert\dot{\phi}(e_p)e_p(t) + \phi(e_p)(t)\dot{e}_p(t)\Vert^2\nonumber \\ 
&-\frac{K_{LK_{2}}\omega_2}{2} \int_{t_T}^{t}[\frac{d}{dt}(\phi(e_p)(l) e_p(l))]^T [\frac{d}{dt}(\phi(e_p)(l) e_p(l))] dl \nonumber
\end{align}
where $\phi(e_p)$, and $\dot{\phi}(e_p)$ denote diagonal matrices containing terms $\phi(e_{pi})$, and $\dot{\phi}(e_{pi}), \;\forall i=\{1,..,n\}$, respectively. Substituting for $\dot{\phi}$, $\dot{\eta}$ and $\dot{e}_p$, 
using triangle and Young's inequalities, the following upper bound on the first term of $\dot{P}_2$ is obtained
\begin{align}
&k_1^26K_{LK_{2}}\omega_2T \Vert\phi^3(e_p) e_p^2\Vert^2\Vert e_p \Vert^2 \nonumber \\ 
&+6K_{LK_{2}}\omega_2Tw_j^2 \Vert \phi^2(e_p)e_p^2 \Vert^2\Vert \dot{\theta}\Vert^2 \Vert \tilde{\zeta}_{j}\Vert^2 \nonumber \\ 
&+6K_{LK_{2}}\omega_2Tj_p^2\Vert \phi^2(e_p)e_p^2\Vert^2\Vert\eta\Vert^2 \nonumber \\
&+ k_1^26K_{LK_{2}}\omega_2T\Vert \phi^2(e_p) e_p\Vert^2 \nonumber \\ 
&+6K_{LK_{2}}\omega_2Tw_j^2\Vert \phi(e_p)\Vert^2\Vert \dot{\theta}\Vert^2 \Vert \tilde{\zeta}_{j}\Vert^2 \nonumber \\ 
&+6K_{LK_{2}}\omega_2 Tj_p^2\Vert\phi(e_p)\Vert^2 \Vert\eta\Vert^2\leq T\rho_1(\Vert y \Vert)\Vert y \Vert^2
\end{align}
where $y=[e_p^T\; \eta^T \; \tilde{\zeta}_j^T]^T$, $\rho_1(\Vert y \Vert)$ is a positive invertible non-decreasing function. The symbol $\phi^3(e_p)e_p^2 = [\phi^3(e_{p1})e_{p1}^2,...,\phi^3(e_{pn})e_{pn}^2]^T$ is a vector containing elements $\phi^3(e_{pi})e_{pi}^2$, $\phi^2(e_{pi})e_{pi}^2 = [\phi^2(e_{p1})e_{p1}^2,...,\phi^2(e_{pn})e_{pn}^2]^T$ is a vector containing elements of $\phi^2(e_{pi})e_{pi}^2$, and $\phi^2(e_p)e_p$ is a vector containing elements $\phi^2(e_{pi})e_{pi}$. 

$\dot{P}_3$ is computed as
\begin{equation}
\dot{P}_3 \leq Tc_2-\frac{K_{LK_{3}}\omega_3}{2}\int_{t_T}^{t} \dot{p}_h^T(l)\dot{p}_h(l) dl\nonumber
\end{equation}
where $c_2 \geq \frac{K_{LK_{3}}\omega_3}{2}\Vert \dot{p}_h(t) \Vert^2 \in \mathbb{R}^+$ is a constant.

\section{Development of a bound on $\eta^T (W_y\zeta - W_{yT}\zeta)$} \label{app:Appendix2}
Consider following term from (34) (of the main paper) 
\begin{align}
\eta^T(W_y &- W_{y_{T}})\zeta_y = M[  \dot{\hat{J}}_{p}^{-1}
(\dot{p}_{h}-\dot{p}_{h_{T}}) \nonumber \\
&+ k_1\dot{\hat{J}}_{p}^{-1} (\phi(e_p) e_p-\phi(e_{p_{T}}) e_{p_{T}}) \nonumber \\
&+ 
\hat{J}_{p}^{-1}(\ddot{p}_{h} - \ddot{p}_{h_{T}}) + k_1 \hat{J}_{p}^{-1}(\dot{\phi}(e_p) e_p \nonumber \\
&-\dot{\phi}(e_{p_{T}}) e_{p_{T}}) + k_1 \hat{J}_{p}^{-1}(\phi(e_p) \dot{e}_p - \phi(e_{p_{T}}) \dot{e}_{p_{T}})] \nonumber \\
&+ C(\theta,\dot{\theta})(\eta - \eta_T) \label{eq:WzetaBound}
\end{align}
The norm bound can be written as
\begin{align}
&\Vert \eta \Vert \Vert(W_y - W_{y_{T}})\zeta_y \Vert\leq   \bar{m}  \bar{j}_{p_{inv}}^2 \bar{j}_{p_{der}}\Vert \dot{\theta} \Vert \Vert \eta \Vert
\Vert \int_{t_T}^{t} \ddot{p}_h(l) dl\Vert   \nonumber \\
&+ k_1\bar{m}  \bar{j}_{p_{inv}}^2 \bar{j}_{p_{der}}\Vert \dot{\theta} \Vert\Vert \eta \Vert \Vert\int_{t_T}^{t} \frac{d}{dl}\phi(e_p)(l) e_p(l) dl\Vert \nonumber \\
&+ k_1 \bar{m}\bar{j}_{p_{inv}}\Vert \eta \Vert \Vert \dot{\phi}(e_p) e_p -\dot{\phi}(e_{p_{T}}) e_{p_{T}} + \phi(e_p) \dot{e}_p  \nonumber \\
&- \phi(e_{p_{T}}) \dot{e}_{p_{T}} \Vert +\bar{c} \Vert\dot{\theta}\Vert\Vert \eta \Vert \Vert \eta - \eta_T \Vert+ 
\bar{m}\bar{j}_{p_{inv}}\bar{p}_3\Vert \eta \Vert
\end{align}
where $\Vert C(\theta,\dot{\theta}) \Vert \leq \bar{c}\Vert \dot{\theta} \Vert$ is used. Using Young's inequality on select terms, and the following bound on $\Vert\eta-\eta_T\Vert$ developed using (7) and (9) (of the main paper) 
\begin{equation}
\Vert \eta - \eta_T \Vert^2  \leq 2j_{p_{i}}^2 
\Vert \int_{t_T}^{t} \ddot{p}_h dl \Vert ^2 + 2k_1^2 j_{p_{i}}^2\Vert \int_{t_T}^{t}\frac{d}{dl}(\phi(e_p) e_p dl\Vert ^2 \nonumber
\end{equation}
the following upper bound can be developed
\begin{align}
&\Vert \eta \Vert \Vert(W_y - W_{y_{T}})\zeta_y \Vert\leq  
  \bar{m}\bar{j}_{p_{inv}}\bar{p}_3 
\Vert \eta \Vert \nonumber \\
&+ (\frac{(k_1+1)\bar{m}  \bar{j}_{p_{inv}}^2 \bar{j}_{p_{der}}\gamma_7^2+\bar{c}\gamma_9^2}{4}\Vert \dot{\theta} \Vert^2 )\Vert \eta \Vert^2 \nonumber \\
&+ (\frac{\bar{m}  \bar{j}_{p_{inv}}^2 \bar{j}_{p_{der}}}{\gamma_7^2}+\frac{\bar{c}2j_{p_{i}}^2}{\gamma_9^2})  
\Vert \int_{t_T}^{t} \ddot{p}_h(l) dl \Vert ^2\nonumber \\
&+ k_1 \bar{m}\bar{j}_{p_{inv}}  \Vert \eta \Vert \Vert \dot{\phi}(e_p) e_p -\dot{\phi}(e_{p_{T}}) e_{p_{T}} \nonumber \\
&+ \phi(e_p) \dot{e}_p - \phi(e_{p_{T}}) \dot{e}_{p_{T}} \Vert \nonumber \\
&+ (\frac{k_1\bar{m}  \bar{j}_{p_{inv}}^2 \bar{j}_{p_{der}}}{\gamma_7^2} +\frac{\bar{c} 2k_1^2 j_{p_{i}}^2}{\gamma_9^2}) \Vert\int_{t_T}^{t} \frac{d}{dl} \phi(e_p)(l) e_p(l) dl\Vert^2  \label{eq:BoundEtaWZeta}
\end{align}
To further develop the bound in (\ref{eq:BoundEtaWZeta}), following bounds are developed. 
Using the triangle inequality of vector norms results in
\begin{align}
\Vert \dot{\phi}(e_p) e_p -&\dot{\phi}(e_{p_{T}}) e_{p_{T}} \Vert \leq \Vert 2k_1(\phi^3(e_p)e_p^2 - \phi^3(e_{p_{T}}) e_{p_{T}}^2)\Vert \nonumber \\
& +\Vert 2(\phi^2(e_p)e_p - \phi^2(e_{p_{T}})e_{p_{T}})\Vert \Vert W_j \tilde{\zeta}_{j} \Vert \nonumber \\
&+ \Vert2\phi^2(e_p)e_p\hat{J}_p\eta - 2\phi^2(e_{p_{T}})e_{p_{T}}\hat{J}_p\eta_T\Vert \nonumber
\end{align}
where the symbol $\phi^3(e_p)e_p^2 = [\phi^3(e_{p1})e_{p1}^2,...,\phi^3(e_{pn})e_{pn}^2]^T$ is a vector containing elements $\phi^3(e_{pi})e_{pi}^2$ and $\phi^2(e_p)e_p$ is a diagonal matrix containing elements $\phi^2(e_{pi})e_{pi}$. Using Appendix A of \cite{de1997adaptive}, bounds on each term can be developed as 
\begin{align}
&\Vert 2k_1(\phi^3(e_p)e_p^2 - \phi^3(e_{p_{T}}) e_{p_{T}}^2)\Vert \leq \rho_2(\Vert e_p \Vert) \Vert p_h -p_{h_{T}} \Vert \nonumber \\
&\Vert 2(\phi^2(e_p)e_p - \phi^2(e_{p_{T}})e_{p_{T}})\Vert \leq \rho_3(\Vert e_p \Vert) \Vert p_h -p_{h_{T}} \Vert \nonumber \\
&\Vert2\phi^2(e_p)e_p\hat{J}_p\eta - 2\phi^2(e_{p_{T}})e_{p_{T}}\hat{J}_p\eta_T\Vert \nonumber \\
&\leq \rho_{4}(\Vert z_1 \Vert)\Vert z_1(t) - z_1(t-T) \Vert 
\end{align}
where $\rho_i, \; \forall i = \{2,3,4\}$ are positive invertible functions, $z_1(t)= [e_{p}^T \;\eta^T\;]^T \in \mathcal{D} \subset \mathbb{R}^{2n}$ and $z_1(t) - z_1(t-T) = [(p_h - p_{h_{T}})^T \;(\eta - \eta_T)^T \;]^T$.
\begin{align}
    &\Vert \eta \Vert \Vert \dot{\phi}(e_p) e_p -\dot{\phi}(e_{p_{T}}) e_{p_{T}} \Vert \leq\Vert \eta \Vert \rho_2(\Vert e_p \Vert) \Vert p_h -p_{h_{T}} \Vert  \nonumber \\
    &+\Vert \eta \Vert \rho_3(\Vert e_p \Vert) \Vert p_h -p_{h_{T}} \Vert\Vert \Vert W_j \tilde{\zeta}_{j} \Vert + \rho_{4}(\Vert z_1 \Vert)  \Vert \eta \Vert\Vert \eta - \eta_T \Vert \nonumber \\
    &+ \rho_{4}(\Vert z_1 \Vert)  \Vert \eta \Vert\Vert p_h - p_{hT} \Vert 
\end{align}
Using Young's inequality
\begin{align}
    &\Vert \eta \Vert \Vert \dot{\phi}(e_p) e_p -\dot{\phi}(e_{p_{T}}) e_{p_{T}} \Vert \leq\rho_2(\Vert e_p \Vert)\Vert \eta \Vert \Vert p_h -p_{h_{T}} \Vert  \nonumber \\
    &+ \rho_{4}(\Vert z_1 \Vert)  \Vert \eta \Vert\Vert p_h - p_{hT} \Vert \nonumber \\
    &+\frac{\bar{p}_1 w_j \rho_3^2(\Vert e_p \Vert) \Vert \dot{\theta}\Vert^2\gamma_{11}^2 + \rho_{4}^2(\Vert z_1 \Vert)\gamma_{12}^2}{4}\Vert \eta \Vert^2 +\frac{\bar{p}_1 w_j }{\gamma_{11}^2} \Vert \tilde{\zeta}_{j} \Vert^2  \nonumber \\
    &+\frac{2j_{p_{i}}^2}{\gamma_{12}^2} 
\Vert \int_{t_T}^{t} \ddot{p}_h(l) dl \Vert ^2 + \frac{2k_1^2 j_{p_{i}}^2}{\gamma_{12}^2}\Vert \int_{t_T}^{t}\frac{d}{dl}(\phi(e_p)(l) e_p(l)) dl\Vert ^2  \label{eq:BoundTerm1}
\end{align}
Consider following term from (\ref{eq:BoundEtaWZeta})
\begin{align}
&\Vert \phi(e_p) \dot{e}_p -\phi(e_{p_{T}}) \dot{e}_{p_{T}} \Vert \leq \Vert k_1(\phi^2(e_p)e_p - \phi^2(e_{p_{T}})e_{p_{T}}) \Vert \nonumber \\
&+\Vert\phi(e_p) - \phi(e_{p_{T}})\Vert \Vert W_j\tilde{\zeta}_j\Vert+ \Vert\phi(e_p)\hat{J}_p \eta - \phi(e_{p_{T}})\hat{J}_p \eta_T\Vert \nonumber
\end{align}
where $\phi(e_p)$ is a diagonal matrix containing terms $\phi(e_{pi})$, $\phi^2(e_p)e_p$ is a vector containing terms $\phi^2(e_{pi})e_{pi}$. Using Appendix A of \cite{de1997adaptive}, each term can be norm bounded as
\begin{align}
&\Vert k_1(\phi^2(e_p)e_p - \phi^2(e_{p_{T}})e_{p_{T}}) \Vert \leq \rho_5(\Vert e_p \Vert) \Vert p_h -p_{h_{T}}\Vert \nonumber \\
&\Vert(\phi(e_p) - \phi(e_{p_{T}}))\Vert \leq \rho_6(\Vert e_p \Vert) \Vert p_h -p_{h_{T}} \Vert \nonumber \\
&\Vert\phi(e_p)\hat{J}_p \eta - \phi(e_{p_{T}})\hat{J}_p \eta_T\Vert \leq \rho_{7}(\Vert z_1 \Vert)\Vert z_1(t) - z_1(t-T) \Vert \nonumber
\end{align}
such that $\rho_i, \; \forall i = \{ 5,6,7\}$ are positive invertible functions. The norm bound can be written as
\begin{align}
&\Vert \eta \Vert\Vert \phi(e_p) \dot{e}_p -\phi(e_{p_{T}}) \dot{e}_{p_{T}} \Vert \leq \rho_5(\Vert e_p \Vert)\Vert \eta \Vert \Vert p_h -p_{h_{T}}\Vert \nonumber\\
&+\bar{p}_1w_j\rho_6(\Vert e_p\Vert)\Vert \dot{\theta}\Vert\Vert \eta \Vert\Vert\tilde{\zeta}_j\Vert + \rho_{7}(\Vert z_1 \Vert)  \Vert \eta \Vert\Vert \eta - \eta_T \Vert \nonumber \\
& + \rho_{7}(\Vert z_1 \Vert)  \Vert \eta \Vert\Vert p_h - p_{hT} \Vert
\end{align}
Using Young's Inequality,
\begin{align}
    &\Vert \eta \Vert\Vert \phi(e_p) \dot{e}_p -\phi(e_{p_{T}}) \dot{e}_{p_{T}} \Vert \leq \rho_5(\Vert e_p \Vert) \Vert \eta \Vert\Vert p_h -p_{h_{T}} \Vert  \nonumber \\
    & + \rho_{7}(\Vert z_1 \Vert)  \Vert \eta \Vert\Vert p_h - p_{hT} \Vert \nonumber \\
    &+\frac{\bar{p}_1 w_j \rho_6^2(\Vert e_p \Vert) \Vert \dot{\theta}\Vert^2\gamma_{13}^2 +\rho_{7}^2(\Vert z_1 \Vert)\gamma_{14}^2}{4}\Vert \eta \Vert^2 +\frac{\bar{p}_1 w_j }{\gamma_{13}^2} \Vert \tilde{\zeta}_{j} \Vert^2  \nonumber \\
    &+\frac{2j_{p_{i}}^2}{\gamma_{14}^2} 
\Vert \int_{t_T}^{t} \ddot{p}_h(l) dl \Vert ^2 + \frac{2k_1^2 j_{p_{i}}^2}{\gamma_{14}^2}\Vert \int_{t_T}^{t}\frac{d}{dl}(\phi(e_p)(l) e_p(l)) dl\Vert ^2  \label{eq:BoundTerm2}
\end{align}
Using (\ref{eq:BoundTerm1}) and (\ref{eq:BoundTerm2}), the bound in (\ref{eq:BoundEtaWZeta}) can be written as
\begin{align}
&\Vert \eta \Vert \Vert(W_y - W_{y_{T}})\zeta_y \Vert\leq  
  \bar{m}\bar{j}_{p_{inv}}\bar{p}_3 
\Vert \eta \Vert + \rho_9(\Vert z_1\Vert, \Vert \dot{\theta}\Vert)\Vert \eta \Vert^2 \nonumber \\
&+ c_{LK_{1}}  
\Vert \int_{t_T}^{t} \ddot{p}_h(l) dl \Vert ^2+ c_{LK_{2}} \Vert\int_{t_T}^{t} \frac{d}{dl} \phi(e_p)(l) e_p(l) dl\Vert^2 \nonumber \\
&+c_{LK_{3}} \Vert \int_{t_{t}}^t \dot{p}_h dl \Vert^2 +(\frac{\bar{p}_1 w_j }{\gamma_{11}^2}+\frac{\bar{p}_1 w_j }{\gamma_{13}^2}) \Vert \tilde{\zeta}_{j} \Vert^2  
\end{align}
where $\rho_8(\Vert z_1 \Vert) = \rho_2(\Vert e_p \Vert)+\rho_5(\Vert e_p \Vert) + \rho_{4}(\Vert z_1 \Vert) + \rho_{7}(\Vert z_1 \Vert) $,$\rho_9(\Vert z_1\Vert, \Vert \dot{\theta}\Vert) = (\frac{(k_1+1)\bar{m}  \bar{j}_{p_{inv}}^2 \bar{j}_{p_{der}}\gamma_7^2+\bar{c}\gamma_9^2+\bar{p}_1 w_j \rho_3^2(\Vert e_p \Vert) \Vert \dot{\theta}\Vert^2\gamma_{11}^2 + \rho_{4}^2(\Vert z_1 \Vert)\gamma_{12}^2}{4}\Vert \dot{\theta} \Vert^2 +\frac{\bar{p}_1 w_j \rho_6^2(\Vert e_p \Vert) \Vert \dot{\theta}\Vert^2\gamma_{13}^2 +\rho_{7}^2(\Vert z_1 \Vert)\gamma_{14}^2}{4} +\frac{\gamma_{15}^2\rho_8^2(\Vert z_1 \Vert)}{4})$,$c_{LK_{1}} = (\frac{\bar{m}  \bar{j}_{p_{inv}}^2 \bar{j}_{p_{der}}}{\gamma_7^2}+\frac{\bar{c}2j_{p_{i}}^2}{\gamma_9^2}+\frac{2j_{p_{i}}^2}{\gamma_{12}^2} 
+\frac{2j_{p_{i}}^2}{\gamma_{14}^2})$, $c_{LK_{2}} = (\frac{k_1\bar{m}  \bar{j}_{p_{inv}}^2 \bar{j}_{p_{der}}}{\gamma_7^2} +\frac{\bar{c} 2k_1^2 j_{p_{i}}^2}{\gamma_9^2} + \frac{2k_1^2 j_{p_{i}}^2}{\gamma_{12}^2} + \frac{2k_1^2 j_{p_{i}}^2}{\gamma_{14}^2})$, $c_{LK_{3}} =\frac{1}{\gamma_{15}^2}$. 

\section{Development of bounds on cross terms} \label{app:Appendix3}
Using Cauchy-Schwarz,
and Young's inequality
and $\Vert W_j \Vert \leq w_j \Vert \dot{\theta} \Vert$
\begin{align}
\Vert\tilde{\zeta}_j^TW_j^T \int \frac{d}{dt} \phi(e_p)e_p dl\Vert &\leq  \frac{\gamma_6^2}{4}w_j^2\Vert \dot{\theta} \Vert^2 \Vert\tilde{\zeta}_j \Vert^2  \nonumber \\ 
&+  \frac{1}{\gamma_6^2}\Vert\int \frac{d}{dt} \phi(e_p)e_p dl \Vert^2
\end{align}
where $\gamma_6 \in \mathbb{R}^+$ is a constant. 
Similarly using Cauchy-Schwarz, 
Young's inequality
and $\Vert W_{y_{T}} \Vert \leq \rho_2(\Vert y_T\Vert) = w_{y_{T}}\Vert y_T\Vert + \bar{w}_{y_{T}}$, where $y_T= [e_{p_{T}}^T \;  \eta_T^T \;\hat{\zeta}_j^T \;]^T \in \mathcal{D} \subset \mathbb{R}^{2n+m_1}$ and $\Vert W_{y} \Vert \leq w_{y}\Vert y\Vert + \bar{w}_{y}$, we have
\begin{align}
\Vert\tilde{\zeta}_y^T W_{y_{T}}^T (\eta -\eta_T)\Vert \leq& \frac{\gamma_5^2}{4}\Vert\tilde{\zeta}_y\Vert^2  + \frac{2j_{p_{i}}^2}{\gamma_5^2}\rho_{10}^2(\Vert y_T\Vert) 
\nonumber \\& \Vert \int_{t_T}^{t} \ddot{p}_h(l) dl \Vert ^2\!+\! \frac{ 2k_1^2 j_{p_{i}}^2}{\gamma_5^2}\rho_{10}^2(\Vert y_T\Vert)\nonumber \\ 
&\Vert \int_{t_T}^{t}\frac{d}{dt}(\phi(e_p)(l) e_p(l)) dl\Vert ^2
\end{align}

\bibliographystyle{IEEEtran}
\phantomsection\addcontentsline{to}{section}{\refname}\bibliography{RCL_Complete}
\end{document}